\documentclass{article}
\pdfoutput=1

\usepackage[numbers, compress]{natbib}
\usepackage[preprint]{neurips_2022}

\usepackage[utf8]{inputenc} 
\usepackage[T1]{fontenc}    
\usepackage{hyperref}       
\usepackage{url}            
\usepackage{booktabs}       
\usepackage{amsfonts}       
\usepackage{nicefrac}       
\usepackage{microtype}      
\usepackage{xcolor}         
\usepackage{amsmath}
\usepackage{amssymb}
\usepackage{graphicx}
\usepackage{wrapfig}
\usepackage{multirow}
\usepackage{multicol}
\usepackage{tabularx}
\usepackage{subfig}
\usepackage{float}

\usepackage{pifont}
\newcommand{\cmark}{\ding{51}}%

\title{On Data Scaling in Masked Image Modeling}

\author{%
  Zhenda Xie\thanks{~Equal Contribution. The work is done when Zhenda Xie, Yutong Lin, and Yixuan Wei are interns at Microsoft Research Asia. }\hspace{1.2mm}$^{13}$, Zheng Zhang$^{*3}$, Yue Cao$^{*3}$, Yutong Lin$^{23}$, Yixuan Wei$^{13}$, Qi Dai$^{3}$, Han Hu$^{3}$ \\
  $^1$Tsinghua University\\
  $^2$Xi'an Jiaotong University\\
  $^3$Microsoft Research Asia\\
}

\begin{document}

\maketitle

\begin{abstract}
An important goal of self-supervised learning is to enable model pre-training to benefit from almost unlimited data. However, one method that has recently become popular, namely masked image modeling (MIM), is suspected to be unable to benefit from larger data. In this work, we break this misconception through extensive experiments, with data scales ranging from 10\% of ImageNet-1K to full ImageNet-22K, model sizes ranging from 49 million to 1 billion, and training lengths ranging from 125K iterations to 500K iterations. Our study reveals that: (i) Masked image modeling is also demanding on larger data. We observed that very large models got over-fitted with relatively small data; (ii) The length of training matters. Large models trained with masked image modeling can benefit from more data with longer training; (iii) The validation loss in pre-training is a good indicator to measure how well the model performs for fine-tuning on multiple tasks. This observation allows us to pre-evaluate pre-trained models in advance without having to make costly trial-and-error assessments of downstream tasks. We hope that our findings will advance the understanding of masked image modeling in terms of scaling ability. 
\end{abstract}

\section{Introduction}
In natural language processing, scaling model capacity and data size has been an important driving force for the remarkable improvements of language models over the past few years~\cite{kaplan2020scaling,radford2019language,raffel2019t5,Turing-17B,brown2020language,fedus2021switch,Megatron-Turing-530B}. Behind the success is a self-supervised pre-training approach, masked language modeling (MLM)~\cite{devlin2018bert}, that can take advantage of and benefit from almost unlimited data. As the same time, the relevant research in the field of computer vision has also been intensifying. However, due to the lack of effective self-supervision methods, most previous works are based on image classification tasks~\cite{tan2019efficientnet,kolesnikov2020big,zhai2021scaling,dai2021coatnet}, where the huge labeling cost and low information contained in the labels limit broader exploration of scaling visual models, or the models being scaled up further, thus leaving progress in computer vision largely behind the NLP field.

Recently, a self-supervision visual pre-training method named masked image modeling (MIM)~\cite{bao2021beit, he2021masked, xie2021simmim} has become popular due to its impressive fine-tuning performance on a variety of downstream computer vision tasks. Given its high analogy with MLM~\cite{devlin2018bert}, the dominant pre-training approach in NLP, we expect masked image modeling to advance the scaling performance of visual models. Specifically, we are concerned with two aspects of scaling ability: model scaling and data scaling. While the masked image modeling approach is shown to be good at scaling up model capacity~\cite{he2021masked,swinv2}, like NLP models, its ability to benefit from larger data is unclear or even a bit negative. For example, \cite{el2021large, tong2022videomae} show that using a small amount of training data in masked image modeling can achieve comparable performance than that using larger data. The data scaling capability is critical, as an important hallmark of self-supervised learning is the ability to leverage almost unlimited data, and failure to benefit from larger data may hinder the future potential of the masked image modeling.

\begin{figure}[t]
    \centering
    \includegraphics[width=1\linewidth]{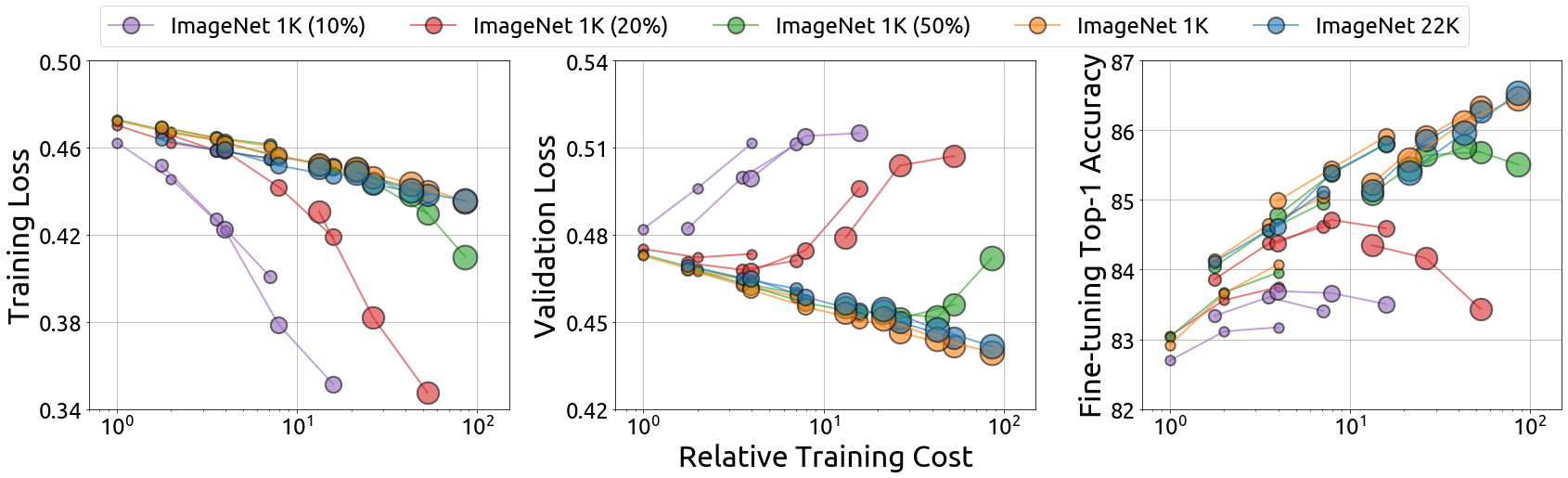}
    \caption{The curves of training loss, validation loss of pre-training, and fine-tuning accuracy on ImageNet-1K of different model sizes, data sizes and training lengths, w.r.t. the relative training cost. We set the training cost of SwinV2-S for 125K iterations as the value of 1. Bigger circles indicate larger models. \emph{Best viewed in color.}
    }
    \label{fig:teaser}
\end{figure}

In this paper, we systematically investigate the data scaling capability of masked image modeling at different model sizes and training lengths. We use Swin Transformer V2~\cite{swinv2} as the visual encoder because of its proven trainability for large models and its applicability to a wide range of vision tasks, and adopt SimMIM~\cite{xie2021simmim} for masked image modeling pre-training because it has no restrictions on the encoder architectures. With extensive experiments, we find that:

(i) \textit{Masked image modeling is demanding for large data.} We observed large models overfited with relatively small data, as reflected by the increased validation losses with longer training when a large model while relatively small data is used (see Figure~\ref{fig:teaser} center). The overfitting issue will result in degraded fine-tuning performance, as shown in Figure~\ref{fig:teaser} right.

(ii) \textit{Training length matters. Large models trained with masked image modeling can benefit from more data at a longer training length.} When the training length is short, the difference in performance between using large and small datasets is not significant. However, with sufficient training, more data shows better performance. In addition, as the data size increases, the fine-tuning performance of large models saturates more slowly than that of small models.

(iii) \textit{The validation loss is a good proxy indicator for fine-tuning performance.} We observe a strong correlation between validation loss and fine-tuning performance on multiple tasks. This finding suggests that the validation loss can be used as a good indicator of how well the model is trained, which can reduce the overhead of evaluation by direct fine-tuning on downstream tasks.

These findings suggest that masked image modeling (MIM) is not only a \textit{model} scalable learner, but also a \textit{data} scalable learner. Particularly, our revealing of data scaling capability of masked image modeling breaks the misconception of previous studies that suspected masked image modeling could not benefit from more data. We hope these findings will deepen the understanding of masked image modeling.

\section{Background and Experimental Setup}

\subsection{Masked Image Modeling}
Masked image modeling is used to train the vision model by taking a corrupted image as input and predicting the content of the masked region as the target. In this study, we use SimMIM~\cite{xie2021simmim} as the default masked image modeling approach because of its simplicity and lack of restrictions on the architecture of the vision encoder. SimMIM consists of a visual encoder and an extremely lightweight prediction head of a linear layer for predicting the raw pixels of the corrupted images via $\ell_2$ regression loss. To facilitate the implementation of the vision transformer, SimMIM adopts the patch-wise mask strategy with the masked patch size of $32\times32$ and mask ratio of $0.6$. To further alleviate the local dependency of raw pixels, we improved the SimMIM by normalizing the predicted target according to~\cite{fang2022corrupted} with a sliding window of $47^2$. As the result, a slight performance improvement is observed.

\begin{table}[h]\small
    \centering
    \begin{tabular}{c|c|c|c|c|c|c}
    \toprule
    \multirow{2}{*}{Model} & {Base} & \multirow{2}{*}{Depth} & \multirow{2}{*}{Head} & \multicolumn{2}{c|}{Window Size} & Backbone \\
    \cline{5-6}
    & Channel & & & pre-train & fine-tune & Params \\
    \hline
    SwinV2-S & 96 & \{2, 2, 18, 2\} &  \{3, 6, 12, 24\} & 12 & 14 & 49M\\
    SwinV2-B & 128 & \{2, 2, 18, 2\} &  \{4, 8, 16, 32\} & 12 & 14 & 87M\\
    SwinV2-L & 192 & \{2, 2, 18, 2\} &  \{6, 12, 24, 48\} & 12 & 14 & 195M\\
    SwinV2-H & 352 & \{2, 2, 18, 2\} &  \{11, 22, 44, 88\} & 12 & 14 & 655M\\
    SwinV2-g & 448 & \{2, 2, 18, 2\} &  \{14, 28, 56, 112\} & 12 & 14 & 1061M\\
    \hline
    \end{tabular}
    \vspace{0.5em}
    \caption{Detailed architecture specifications.  SwinV2-g (giant) is a new variant to those in~\cite{swinv2}, with number of parameters between SwinV2-L and the 3-billion-parameter SwinV2-G (Giant). }
    \label{table:Swin_variants}
\end{table}
\subsection{Architecture Specifications}
We use Swin Transformer V2~\cite{swinv2} as the vision encoder in this study. Thanks to its generality and scalability, we evaluate a series of SwinV2 models with a wide range of model sizes (the number of parameters ranges from $\sim$50M to $\sim$1B, and FLOPs range from $\sim$9G to $\sim$190G) on multiple downstream tasks. The detailed model specifications are shown in Table~\ref{table:Swin_variants}. We use a new variant SwinV2-g (giant), with number of parameters between SwinV2-L and the 3-billion-parameter SwinV2-G (Giant) used in~\cite{swinv2}.

\begin{table}[t]\small
    \centering
    \begin{tabular}{c|c|c|c|c|c|c}
    \toprule
    & IN1K ($10\%$) & IN1K ($20\%$) & IN1K ($50\%$) & IN100 &IN1K ($100\%$) & IN22K ($100\%$) \\
    \hline
    \# Classes & $1\times10^3$ & $1\times10^3$ & $1\times10^3$ & $1\times10^2$ & $1\times10^3$ & $2.18 \times 10^4$\\
    \# Images & $1.28\times10^5$ & $2.56\times10^5$ & $6.41\times10^5$ & $1.27\times10^5$ & $1.28\times10^6$ & $1.42 \times 10^7$ \\
    \hline
    \end{tabular}
    \vspace{0.5em}
    \caption{Detailed dataset specifications used in the pre-training of masked image modeling.}
    \label{table:pre_train_dataset}
\end{table}
\subsection{Pre-training Datasets}
To study the effect of data size on masked image modeling, we build datasets with different sizes. We use the training set of ImageNet-1K and ImageNet-22K as two large-scale datasets, and randomly sample $10\%$, $20\%$, $50\%$ of images in the ImageNet-1K training set as smaller datasets. By default, the images are uniformly sampled from each category. We also consider the sampling strategies could perform differently. To this end, we randomly sample 100 classes from ImageNet-1K as ImageNet-100, and compare it with ImageNet-1K ($10\%)$ but find their training loss and fine-tuning performance are almost the same. The details and statistics of all pre-training datasets used in our study are shown in Table~\ref{table:pre_train_dataset}.

\subsection{Pre-training Details}
To better compare the performance of models with different amounts of data under the same pre-training length, we use training iterations rather than training epochs and adopt the same hyper-parameters for all models with different sizes during pre-training. The total number of training iterations is in \{125K, 250K, 500K\} and the batch size is set as 2048 for all experiments. In pre-training stage, we use the same hyper-parameters for all models, and the training details and hyper-parameters of pre-training are summarized in Table~\ref{table:setting-pretrain}. Because of the excessive amount of experiments, we follow SimMIM~\cite{xie2021simmim} and also use the following two techniques for reducing the experimental overheads: First, we use the step learning rate scheduler in pre-training for sharing the first training step among experiments with different training lengths. The first 7/8 training iterations are the first step and the last 1/8 training iterations are the second step with the learning rate ratio of 0.1 (\emph{i.e.} learning rate is divided by 10 in the second step). Second, we adopt the input image size of $192^2$ and set the window size of $12$. We improve the SimMIM by normalizing the predicted target according to ~\cite{fang2022corrupted} with a sliding window of $47^2$ and observe an improvement of 0.3 on top-1 accuracy of ImageNet-1K for the SwinV2-Large model. The same light data augmentation strategy as SimMIM is used: random resize cropping with a scale range of [0.67, 1], an aspect ratio range of [3/4, 4/3] and a random flipping with probability 0.5.

\subsection{Fine-tuning Tasks}

To extensively and accurately evaluate the performance of pre-trained models under different pre-training schedulers and datasets, a series of diverse and representative tasks including fine-tuning on ImageNet-1K, fine-grained image classification, object detection, instance segmentation, and semantic segmentation are selected for evaluation.

\paragraph{ImageNet-1K}
We follow ~\cite{bao2021beit} to evaluate the quality of learnt representations by fine-tuning the pre-trained models on ImageNet-1K~\cite{deng2009imagenet} image classification task, which is the most commonly used scenario and evaluation criterion for pre-trained models~\cite{he2021masked, xie2021simmim}. The setting details and fine-tuning hyper-parameters for ImageNet-1K image classification are summarized in Table~\ref{table:setting-finetune}. Different from pre-training, We adopt the image size with $224^2$ with window size of 14 in fine-tuning. The AdamW with batch size of 2048, base learning rate of 5e-3, weight decay of 0.05, $\beta_1$ of 0.9 and $\beta_2$ of 0.999 are used, and we adopt cosine learning rate scheduler. As larger models are more prone to overfitting, we fine-tune SwinV2-S/B/L for 100 epochs with 20 warm-up epochs and SwinV2-H/g for 50 epochs with 10 warm-up epochs, and decrease the layer decay as the model size increases. In addition, gradient clipping, stochastic depth, label smoothing and data augmentations (\emph{e.g.} random crop, rand erasing~\cite{zhong2020random}, rand augment~\cite{cubuk2020randaugment}, mixup~\cite{zhang2017mixup}, cutmix~\cite{yun2019cutmix}, \emph{etc.}) are also used by following ~\cite{xie2021simmim}. 

\paragraph{iNaturalist-18}
iNaturalist~\cite{van2018inaturalist} 2018 is a long-tailed fine-grained image classification dataset. The details and fine-tuning hyper-parameters for iNaturalist 2018 are summarized in Table~\ref{table:setting-inat}. As fine-tuning in ImageNet-1K, we also use the input image size of $224^2$, window size of 14 and patch size of 4 in iNaturalist 2018. We fine-tune all models for 100 epochs with 20 warm-up epochs, and set layer decay to 0.8, 0.75 and 0.7 for SwinV2-S/B/L, respectively. The AdamW optimizer with cosine learning rate scheduler, batch size of 2048, base learning rate of 1.6e-2, weight decay of 0.1, $\beta_1$ of 0.9 and $\beta_2$ of 0.999 are used. In addition, we also adopt stochastic depth, label smoothing, gradient clipping and data augmentations in fine-tuning.

\paragraph{COCO Object Detection and Instance Segmentation~\cite{lin2014coco}}
The details and fine-tuning hyper-parameters for COCO dataset are summarized in Table~\ref{table:setting-coco}. We use Mask R-CNN~\cite{Mask-rcnn}\footnote{Our implementation based on MMDetection~\cite{chen2019mmdetection}.} for evaluation. We set the window size to 14 and patch size to 4. The AdamW optimizer with batch size of 32, base learning rate of 8e-5, weight decay of 0.05, $\beta_1$ of 0.9, $\beta_2$ of 0.999 and a step learning rate scheduler (step learning rate ratio of 0.1, step epochs are 27 and 33) are used. In training, the random cropping with crop size of [1024, 1024], large scale jittering with a range of [0.1, 2.0], random horizontal flip with probability 0.5, and stochastic depth regularization are used. In testing, all images are resized to (800, 1333) and keeping the aspect ratio unchanged.

\paragraph{ADE20K Semantic Segmentation~\cite{zhou2018ade}} 
The details and fine-tuning hyper-parameters for ADE20K dataset are summarized in Table~\ref{table:setting-ade}. Following \cite{liu2021swin}, we use UPerNet~\cite{xiao2018upernet} for evaluation. We set the window size to 20 and the patch size to 4. The AdamW optimizer with with batch size of 32, base learning rate searched in a range of [1e-4, 3e-4], weight decay of 0.05, $\beta_1$ of 0.9, $\beta_2$ of 0.999 and a linear learning rate scheduler with a total of 80K iterations are used. Also, we use the layer decay of 0.95, 0.95, 0.9 for SwinV2-S/B/L, respectively. In training, the random cropping with crop size of [640, 640], scale jittering with a range of [0.5, 2.0], random horizontal flip with probability 0.5, random photometric distortion and stochastic depth regularization of 0.1 are used. In testing, all images are evaluated by sliding window manner, and use the test image size of (2560, 640) and set sliding window stride to 426, following~\cite{liu2021swin,xie2021simmim}.

\section{Results and Findings}

\begin{figure}[t]
    \centering
    \includegraphics[width=1.0\linewidth]{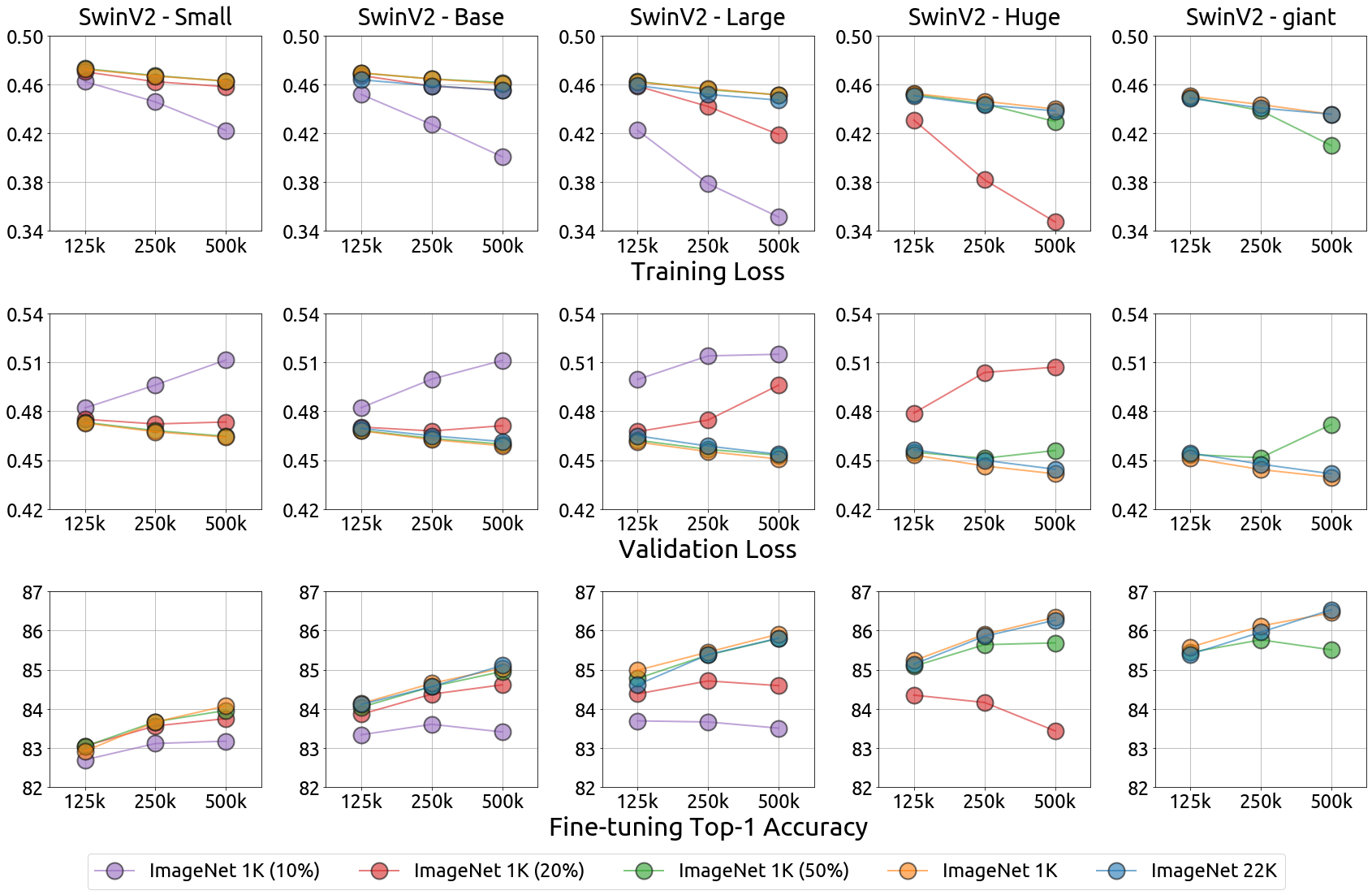}
    \caption{Relationship among training loss, validation loss of pre-training, and fine-tuning performance of ImageNet-1K measured by top-1 accuracy, w.r.t. the training length. \emph{Best viewed in color.}}
    \label{fig:summary}
\end{figure}

\subsection{Training Length, Data Size and Model Size}
We train numerous models with different training lengths, data sizes, and model sizes, and study how these factors affect the performance of masked image modeling. Figure~\ref{fig:teaser} and Figure~\ref{fig:summary} illustrate the relationship between the training loss, and the validation loss of pre-training\footnote{The validation loss of pre-training is measured on the validation set of ImageNet-1K for all experiments.}, and the fine-tuning top-1 accuracy of ImageNet-1K. Based on these extensive experiments, we make the following observations:

\begin{figure}[t]
    \centering
    \includegraphics[width=0.95\linewidth]{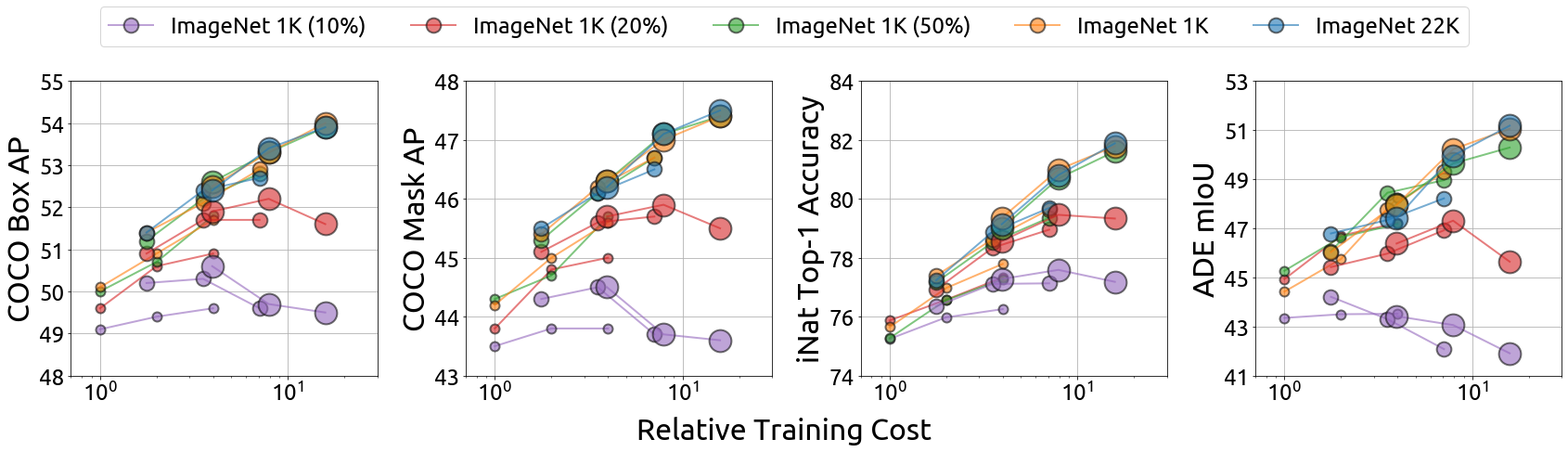}
    \caption{The curves of performances on COCO object detection (a), COCO instance segmentation (b), iNaturalist-18 (c), and ADE20K semantic segmentation (d) w.r.t. the relative training cost. Note that the training cost indicates the pre-training cost. We set the training cost of SwinV2-S for 125K iterations as 1. Bigger circles indicate larger models. \emph{Best viewed in color.}
    }
    \label{fig:teaser_downstream}
\end{figure}

\paragraph{Masked image modeling remains demanding for large dataset.} When with the high masking rate (\emph{e.g.}, $60\%$ in our work), the masked image modeling is considered a very challenging training objective and has been found to be data efficient by previous literature~\cite{el2021large,liu2021video}, \emph{i.e.}, a comparable performance can be achieved with small datasets as with large datasets. However, Figure~\ref{fig:teaser} shows that as the training cost increases, the training loss of some models drops significantly, and their validation loss rises significantly, even on using 50\% images of ImageNet-1K (\emph{i.e.}, IN1K (50\%)), indicating the \textit{overfitting} phenomenon exists. And significant decrease to the fine-tuning performance caused by overfitting could be observed in Figure~\ref{fig:summary}. Moreover, we measure the best fine-tuning performance of each model trained by different training schedulers in Table~\ref{table:data_and_model_acc}. We find the large models perform even worse than smaller models when small dataset is used for training. For example, the best top-1 accuracy of SwinV2-H with IN1K ($20\%)$ is 84.4, worse than the best performance of SwinV2-L by 0.3. In addition, by comparing the best performance that can be obtained using different sizes of dataset, we find that using more data results in better performance. These observations suggest that masked image modeling does not alleviate the demands of large dataset.

\begin{table}\small
    \centering
    \begin{tabular}{c|c|c|c|c|c}
    \toprule
    Model & IN1K (10\%) & IN1K (20\%) & IN1K (50\%) & IN1K (100\%) & IN22K (100\%) \\
    \hline
    SwinV2-S & 83.2 & 83.7 & 84.0 & 84.1 & -\\
    SwinV2-B & 83.6 & 84.6 & 85.0 & 85.0 & 85.1\\
    SwinV2-L & 83.7 & 84.7 & 85.8 & 85.9 & 85.8\\
    SwinV2-H & - & 84.4 & 85.7 & 86.3 & 86.3\\
    SwinV2-g & - & - & 85.8 & 86.5 & 86.5 \\
    \bottomrule
    \end{tabular}
    \vspace{0.5em}
    \caption{The best fine-tuning performance (top-1 accuracy) of each model with different scales of data on ImageNet-1K image classification.}
    \label{table:data_and_model_acc}
\end{table}

\paragraph{The training length matters. Larger models can benefit from more data at a longer training length.} 
By comparing the performance of models pre-trained by different data sizes (3rd row of Figure~\ref{fig:summary}), we find that the fine-tuning performance of the large models saturates more slowly with the increasing data size compared to the smaller models. For example, the SwinV2-S model pre-trained on IN1K ($50\%$) has a very similar fine-tuning performance to the model pre-trained on IN1K ($100\%$). In comparison, the performance difference between the SwinV2-H model pre-trained on IN1K ($50\%$) and IN1K ($100\%$) is near 0.5, which is a significant gap for ImageNet-1K classification. 

Furthermore, a comprehensive observation reveals that the improvements from using more data are not significant under short training lengths. For example, while there is a noticeable performance gap between SwinV2-H trained on IN1K ($50\%$) and IN1K ($100\%$) at a training length of 500K iterations, the gap is less than 0.1 at a training length of 125K iterations. This observation suggests that while larger models can benefit from more data, the training length must also increase at the same time.

\begin{table}\small
    \centering\setlength{\tabcolsep}{4.pt}
    \begin{tabular}{c|c|c|c|c|c|c}
    \toprule
    Model & Iter & IN1K (10\%) & IN1K (20\%) & IN1K (50\%) & IN1K (100\%) & IN22K (100\%) \\
    \hline
    \multirow{3}{*}{SwinV2-S} & 125K & 75.2 & 75.9 & 75.3 & 75.6 & - \\
                              & 250K & 76.0 & 76.6 & 76.6 & 77.0 & - \\
                              & 500K & 76.3 & 77.3 & 77.3 & 77.8 & - \\
    \hline
    \multirow{3}{*}{SwinV2-B} & 125K & 76.4 & 76.9 & 77.2 & 77.4 & 77.2 \\
                              & 250K & 77.1 & 78.3 & 78.5 & 78.6 & 78.9 \\
                              & 500K & 77.1 & 79.0 & 79.4 & 79.6 & 79.7 \\
    \hline
    \multirow{3}{*}{SwinV2-L} & 125K & 77.3 & 78.6 & 79.0 & 79.4 & 79.1 \\
                              & 250K & 77.6 & 79.5 & 80.7 & 81.0 & 80.8 \\
                              & 500K & 77.2 & 79.3 & 81.6 & 81.8 & 81.9 \\                              
    \bottomrule
    \end{tabular}
    \vspace{0.5em}
    \caption{Results of top-1 accuracy on iNaturalist-18 fine-grained image classification.}
    \label{table:iNaturalist2018}
\end{table}

\begin{table}\small
    \centering\setlength{\tabcolsep}{4.pt}
    \begin{tabular}{c|c|c|c|c|c|c}
    \toprule
    Model & Iter & IN1K (10\%) & IN1K (20\%) & IN1K (50\%) & IN1K (100\%) & IN22K (100\%) \\
    \hline
    \multirow{3}{*}{SwinV2-S} & 125K & 43.4 & 44.9 & 45.3 & 44.2 & - \\
                              & 250K & 43.5 & 46.7 & 46.6 & 45.8 & - \\
                              & 500K & 43.5 & 47.2 & 47.2 & 48.3 & - \\
    \hline
    \multirow{3}{*}{SwinV2-B} & 125K & 44.2 & 45.4 & 46.1 & 46.0 & 46.8 \\
                              & 250K & 43.3 & 46.0 & 48.5 & 47.7 & 47.3 \\
                              & 500K & 42.1 & 46.9 & 49.0 & 49.3 & 48.2 \\
    \hline
    \multirow{3}{*}{SwinV2-L} & 125K & 43.4 & 46.4 & 48.0 & 48.0 & 47.4 \\
                              & 250K & 43.1 & 47.3 & 49.6 & 50.2 & 50.0 \\
                              & 500K & 41.9 & 45.6 & 50.3 & 51.1 & 51.2 \\                              
    \bottomrule
    \end{tabular}
    \vspace{0.5em}
    \caption{Results (mIoU) on validation set of ADE20K semantic segmentation.}
    \label{table:ade-20k}
\end{table}

\paragraph{Evaluation on more tasks.}
In addition to ImageNet-1K image classification, we also evaluate the MIM pre-trained SwinV2-S, SwinV2-B and SwinV2-L on iNaturalist-18 fine-grained image classification, ADE20K semantic segmentation, and COCO object detection/segmentation.
Figure~\ref{fig:teaser_downstream} shows a similar pattern with ImageNet-1K (Figure~\ref{fig:teaser} (right)) that as the training cost increases, some models have significantly performance drop. In addition, as shown in Table~\ref{table:iNaturalist2018}, \ref{table:ade-20k}, and~\ref{table:COCO}, the smaller models rapidly reach saturation as the amount of data increases, while larger models can continuously benefit from more data after sufficient training. These results suggest that the conclusions drawn on ImageNet-1K are broadly applicable to other vision tasks. 

\begin{table}[t]\small
    \centering\setlength{\tabcolsep}{3.pt}
    \begin{tabular}{c|c|cc|cc|cc|cc|cc}
    \toprule
    \multirow{2}{*}{Model} & \multirow{2}{*}{Iter} & \multicolumn{2}{c|}{IN1K (10\%)} & \multicolumn{2}{c|}{IN1K (20\%)} & \multicolumn{2}{c|}{IN1K (50\%)} & \multicolumn{2}{c|}{IN1K (100\%)} & \multicolumn{2}{c}{IN22K (100\%)} \\
    \cline{3-12}
    & & AP$^\text{box}$ & AP$^\text{mask}$ & AP$^\text{box}$ & AP$^\text{mask}$ & AP$^\text{box}$ & AP$^\text{mask}$ & AP$^\text{box}$ & AP$^\text{mask}$ & AP$^\text{box}$ & AP$^\text{mask}$ \\
    \hline
    \multirow{3}{*}{SwinV2-S} & 125K & 49.1 & 43.5 & 49.6 & 43.8 & 50.0 & 44.3 & 50.1 & 44.2 & - & - \\
                              & 250K & 49.4 & 43.8 & 50.6 & 44.8 & 50.7 & 44.7 & 50.9 & 45.0 & - & -\\
                              & 500K & 49.6 & 43.8 & 50.9 & 44.8 & 51.8 & 45.7 & 51.7 & 45.6 & - & - \\
    \hline
    \multirow{3}{*}{SwinV2-B} & 125K & 50.2 & 44.3 & 50.9 & 45.1 & 51.2 & 45.3 & 51.4 & 45.4 & 51.4 & 45.5 \\
                              & 250K & 50.3 & 44.5 & 51.7 & 45.6 & 52.2 & 46.1 & 52.1 & 46.2 & 52.4 & 46.1 \\
                              & 500K & 49.6 & 43.7 & 51.7 & 45.7 & 52.8 & 46.7 & 52.9 & 46.7 & 52.7 & 46.5 \\
    \hline
    \multirow{3}{*}{SwinV2-L} & 125K & 50.6 & 44.5 & 51.9 & 45.7 & 52.6 & 46.3 & 52.5 & 46.3 & 52.4 & 46.2 \\
                              & 250K & 49.7 & 43.7 & 52.2 & 45.9 & 53.3 & 47.1 & 53.3 & 47.0 & 53.4 & 47.1 \\
                              & 500K & 49.5 & 43.6 & 51.6 & 45.5 & 53.9 & 47.4 & 54.0 & 47.4 & 53.9 & 47.5 \\   
    \bottomrule
    \end{tabular}
    \vspace{0.5em}
    \caption{Results of box/mask AP on the validation set of COCO object detection and instance segmentation.}
    \label{table:COCO}
\end{table}
\begin{figure}[t]
    \centering
    \includegraphics[width=1.0\linewidth]{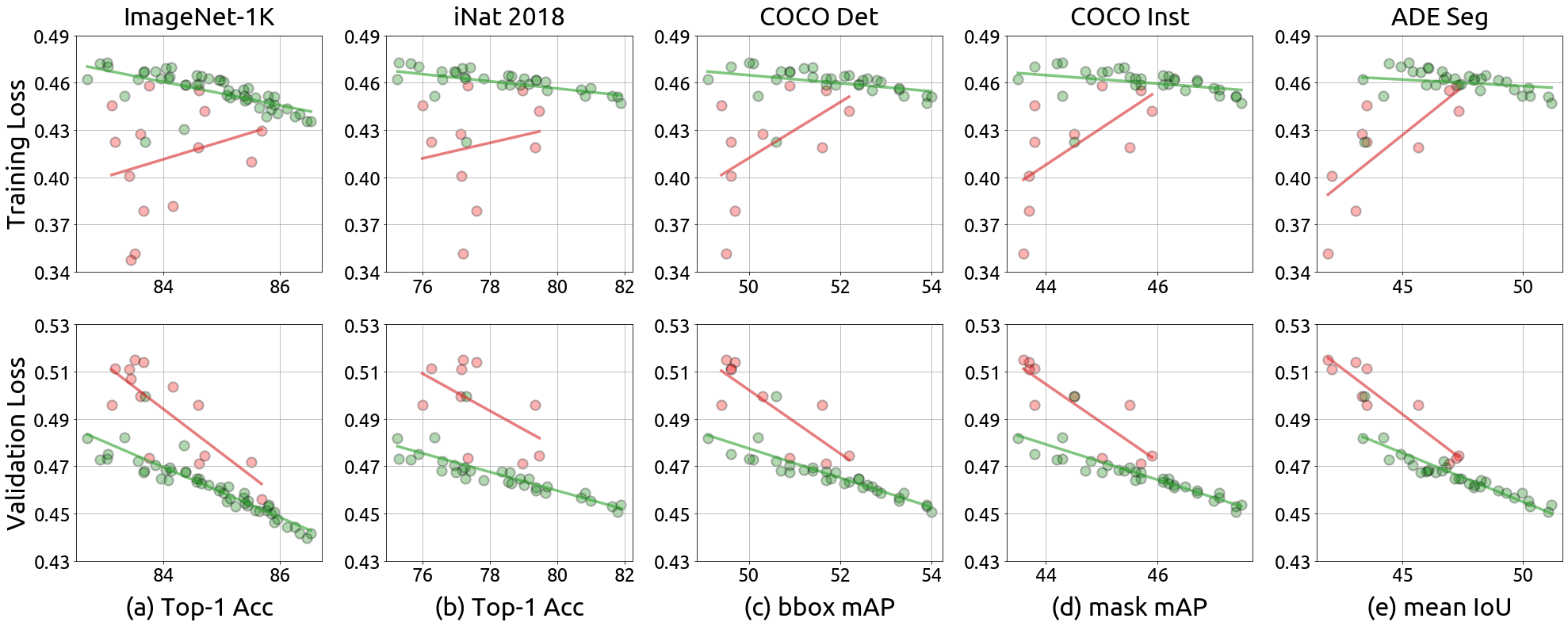}
    \caption{The correlations between pre-training losses (training and validation losses) and the fine-tuning performances. (a) ImageNet-1K image classification; (b) iNat 2018 fine-grained classification; (c) COCO object detection; (d) COCO instance segmentation; (e) ADE20K semantic segmentation. Pre-training losses are highly correlated with fine-tuning performance on all tasks. Red circles are the overfitting models and green circles are non-overfitting models. \emph{Best viewed in color.}}
    \label{fig:corr}
\end{figure}

\begin{figure}[htbp!]
    \centering
    \includegraphics[width=1\linewidth]{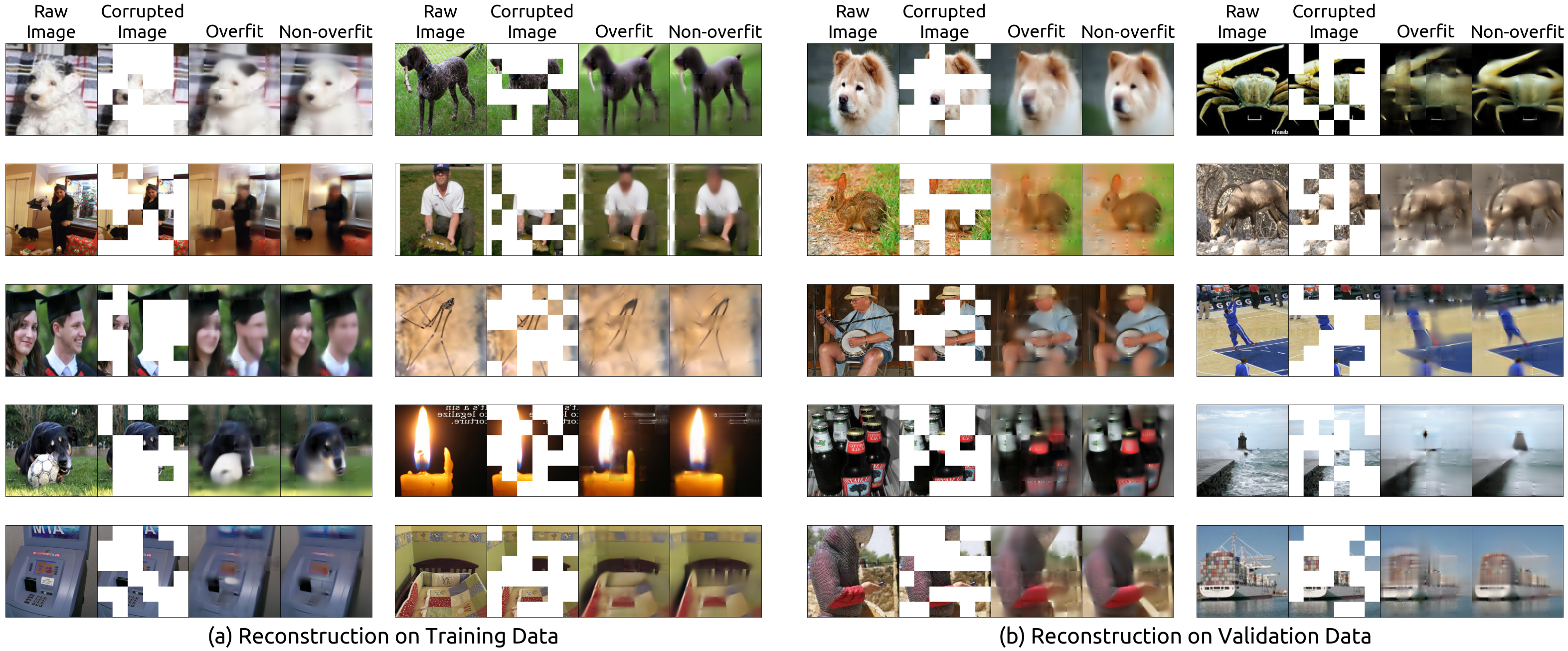}
    \caption{
    We visualize the reconstruction results of overfitting model (SwinV2-L pre-trained on ImageNet-1K(10\%)) and non-overfitting model (SwinV2-L pre-trained on ImageNet-1K(100\%)).
    (a) shows the reconstruction results on the training images from \textbf{ImageNet-1K(10\%)} dataset, which are jointly contained by the training set of two models. (b) shows the reconstruction results on the validation images from \textbf{ImageNet-1K validation set}. Each group contains 4 images from left to right are: the original image, the corrupted images, reconstructed image of overfitting model, and reconstructed image of non-overfitting model.
    }
    \label{fig:recon_train_val}
\end{figure}
\subsection{Reconstruction Results of Overfitting and Non-overfitting Models}
To better understand the difference between overfitting and non-overfitting models, we visualize the reconstruction results of SwinV2-L that pre-trained on ImageNet1K (10\%) and ImageNet1K (100\%). Figure~\ref{fig:recon_train_val}(a) shows the reconstruction results on the training images from ImageNet1K (10\%) dataset, and Figure~\ref{fig:recon_train_val}(b) shows the reconstruction results on the images from ImageNet-1K validation set. Based on the reconstruction results on the training images, we observed the overfitting model (\emph{i.e.} SwinV2-L pre-trained on ImageNet1K (10\%)) is more like "\emph{remembering}" the masked regions, while the non-overfitting model (\emph{i.e.} SwinV2-L pre-trained on ImageNet1K (100\%)) is more like performing "\emph{reasoning}" on the masked regions. For example, the results on the left of the first row in Figure~\ref{fig:recon_train_val}(a) show that the overfitting model even predicts the black hair of the dog, but the seen regions only indicate that the dog is white. And the non-overfitting model only predicts the dog with the white hair.
Furthermore, we observe that the overfitting model seems to lack the "\emph{reasoning}" ability and has a poorer prediction quality on the images of the validation set compared to the non-overfitting model. For example, the results on the left of the first row in Figure~\ref{fig:recon_train_val}(b) show the overfitting model even fails to predict the eyes of the dog. 

\subsection{Correlation between Pre-training Losses and the Fine-tuning Performance}
Evaluating a pre-trained model by its fine-tuned performance on downstream tasks is costly. In supervised pre-training, the validation accuracy is used as the proxy metric to evaluate the quality of the pre-trained models. While in previous studies~\cite{chen2020simclr} on other self-supervised learning approaches (\emph{e.g.}, contrastive learning), such a proxy metric is lacking. In this study, we would like to explore whether the pre-training loss in the training of masked image modeling is a good indicator of its fine-tuning performance. We collect all pre-trained models and plot their training and validation loss curves on Figure~\ref{fig:corr}. Interestingly, the correlations between pre-training losses and the fine-tuning performance on multiple tasks could be observed with a \textit{phase transition} around overfitting. 

Specifically, the correlation between training loss and fine-tuning performance is negative for the overfitting model (green circles) and positive for the non-overfitting model (red circles). The correlation between validation loss and fine-tuning performance is always negative, but the slope of their linear fit lines \footnote{The least squares method is used for linear fit.} is significantly different. 

In addition, we further analyze the Pearson correlation coefficient between training loss and fine-tuning performance (Table~\ref{table:corr}), and find the validation loss has stronger \emph{linear correlation} with fine-tuned performance than train loss for all cases, especially for non-overfitting models.

\subsection{Effects of Different Sizes of Decoders}
We have studied the effects of encoder size from the data scaling perspective. Here, the effects of decoder size are further studied. We pre-train SwinV2-B models with decoder heads of different sizes on IN1K ($20\%$), and Table~\ref{table:diff_heads} shows the results. Interestingly, although we find that the heavier decoder has lower training loss and higher validation loss than the linear decoder, indicating a more severe overfitting issue. But there is no decrease in its fine-tuning performance on ImageNet-1K than the linear decoder. This experiment shows that the decoder behaves very differently from the encoder, and we speculate that this is because the decoder "blocks" the damage to the encoder from overfitting.

\begin{table}\small
    \centering
    \subfloat[ImageNet-1K]{
        \begin{tabular}{c|c|c}
        \toprule
                    & w/ train loss & w/ val loss \\
        \hline
        overfit & +0.26 & -0.79 \\
        non-overfit & -0.64 & -0.90 \\
        \bottomrule
        \end{tabular}
    }
    \subfloat[iNaturalist 2018]{
        \begin{tabular}{c|c|c}
        \toprule
                    &  w/ train loss &  w/ val loss \\
        \hline
        overfit & +0.17 & -0.54 \\
        non-overfit & -0.46 & -0.78 \\
        \bottomrule
        \end{tabular}
    }
    \quad
    \subfloat[COCO Object Detection]{
        \begin{tabular}{c|c|c}
        \toprule
                    &  w/ train loss &  w/ val loss \\
        \hline
        overfit & +0.54 & -0.81 \\
        non-overfit & -0.35 & -0.83 \\
        \bottomrule
        \end{tabular}
    }
    \subfloat[COCO Instance Segmentation]{
        \begin{tabular}{c|c|c}
        \toprule
                    &  w/ train loss &  w/ val loss \\
        \hline
        overfit & +0.62 & -0.86 \\
        non-overfit & -0.31 & -0.85 \\
        \bottomrule
        \end{tabular}
    }
    \quad
    \subfloat[ADE-20K Semantic Segmentation]{
        \begin{tabular}{c|c|c}
        \toprule
                    &  w/ train loss &  w/ val loss \\
        \hline
        overfit & +0.75 & -0.91 \\
        non-overfit & -0.14 & -0.90 \\
        \bottomrule
        \end{tabular}
    }
    \vspace{0.5em}
    \caption{Pearson correlation coefficients between pre-training losses (training and validation losses) and fine-tuning performances on five downstream tasks.}
    \label{table:corr}
\end{table}

\begin{table}\small
    \centering\setlength{\tabcolsep}{7.8pt}
    \begin{tabular}{c|c|c|c|c|c}
    \toprule
    Encoder & Decoder & \# Params & Training loss & Validation loss & Top-1 accuracy \\
    \hline
    SwinV2-B & linear & 90.0M & 0.46 & 0.47 & 84.4 \\
    SwinV2-B & 4-blocks & 140.4M & 0.44 & 0.48 & 84.4 \\
    SwinV2-B & 8-blocks & 190.8M & 0.41 & 0.50 & 84.5\\
    \bottomrule
    \end{tabular}
    \vspace{0.5em}
    \caption{Results of different decoders, including converged training and validation losses of MIM pre-training, and fine-tuning performance (top-1 accuracy) on ImageNet-1K image classification.}
    \label{table:diff_heads}
\end{table}

\subsection{Impact of Different Dataset Sampling Strategies}
\begin{figure}[t]
    \centering
    \includegraphics[width=1.0\linewidth]{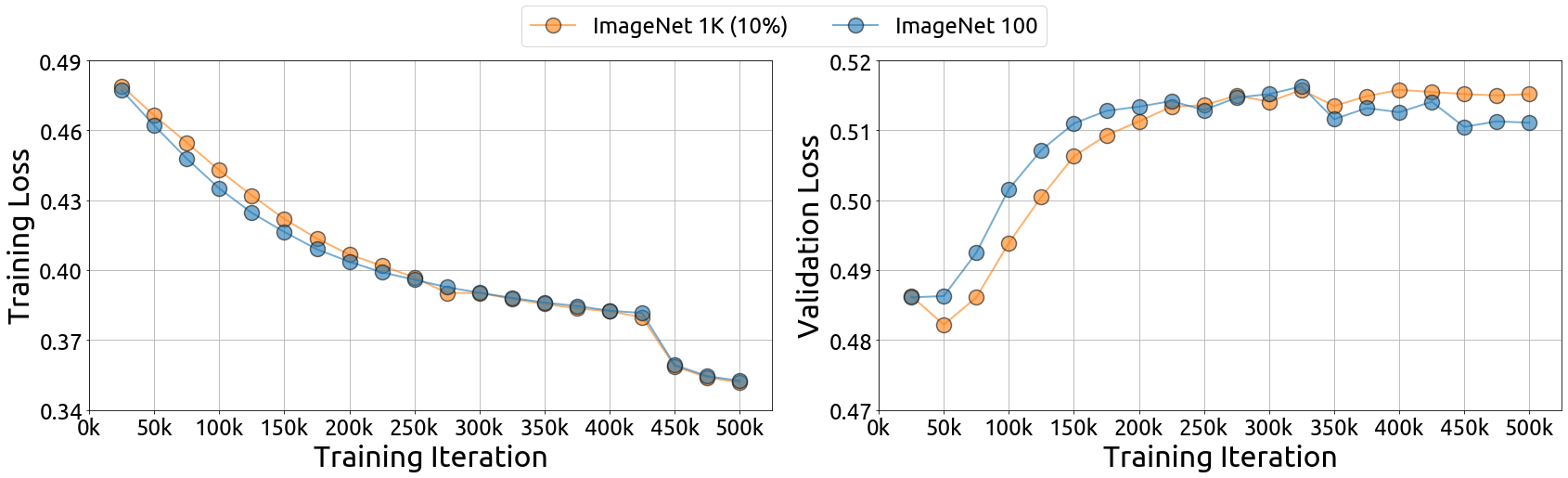}
    \caption{The training loss and validation loss of MIM pre-training with different dataset sampling strategies, ImageNet-1K (10\%) and ImageNet-100. \emph{Best viewed in color.}}
    \label{fig:diff_sampling_curve}
\end{figure}

We study different dataset sampling strategies by comparing the training behavior and fine-tuned performance of models pre-trained on IN1K ($10\%$) and IN100. In IN1K ($10\%$), the images are uniformly sampled from each category, and we randomly sample 100 categories from ImageNet-1K as IN100. Experiments are conducted on SwinV2-L with 500K training iterations. Table~\ref{table:diff_sampling} shows the training loss, validation loss and fine-tuning top-1 accuracy of ImageNet-1K. For the two models pre-trained on IN1K ($10\%$) and IN100, all three metrics are very similar. Figure~\ref{fig:diff_sampling_curve} further illustrates the training dynamics of the two models, and we find both their training loss curves and validation loss curves are almost overlapping. These results show the disparity caused by different dataset sampling strategies is minor.

\section{Related Work}
\paragraph{Masked Image Modeling}
Masked Image Modeling learns representations by reconstructing the masked content of images, and its early exploration can be traced back to context encoder~\cite{pathak2016context} and denoising autoencoder~\cite{vincent2008extracting}. Recently, iGPT~\cite{chen2020imagegpt}, BEiT~\cite{bao2021beit}, MAE~\cite{he2021masked} and SimMIM~\cite{xie2021simmim} recall this approach on training vision transformer. iGPT~\cite{chen2020imagegpt} sequentially predicted the pixels by auto-regressive manner. BEiT~\cite{bao2021beit} proposed to predict the discrete visual tokens. MAE~\cite{he2021masked} and SimMIM~\cite{xie2021simmim} concurrently find predicting the raw pixels with a high masking ratio can work well. In this work, we use SimMIM as the default masked image modeling approach, because of its simplicity and no restrictions on the architecture of vision encoder like MAE.

\paragraph{Vision Transformer}
Transformer~\cite{vaswani2017attention} was first applied to natural language processing and became the dominant architecture, and has recently attracted a lot of attention in computer vision. The pioneering work ViT~\cite{dosovitskiy2020vit} first shows that the transformer architecture works well in image classification when trained on large amounts of data. DeiT~\cite{touvron2021training} proposed a better training recipe based on ViT and demonstrated that vision Transformer has promising performance when only using ImageNet-1K dataset. Swin Transformer~\cite{liu2021swin} improves plain ViT by inducing the hierarchical architecture and non-overlapping local attention and successfully demonstrates the effectiveness of vision transformer on a wide range of vision tasks. Swin Transformer V2~\cite{swinv2} further addresses the training stability issue of ~\cite{liu2021swin} in model scaling and illustrates better performance than the original Swin Transformer, and thus we use it as the default vision encoder in this work.


\begin{table}\small
    \centering
    \begin{tabular}{c|c|c|c|c|c}
    \toprule
    Dataset & \# Images& \# Classes & Training loss & Validation loss & Top-1 Accuracy \\
    \hline
    IN1K ($10\%$) & $1.28\times10^5$ & 1000 & 0.351 & 0.515 & 83.5 \\
    IN100 & $1.27\times10^5$ & 100 & 0.352 & 0.511 & 83.4 \\
    \bottomrule
    \end{tabular}
    \vspace{0.5em}
    \caption{Results on different dataset sampling strategies (ImageNet-1K (10\%) and ImageNet-100), include converged training and validation losses of MIM pre-training, and fine-tuning performance (top-1 accuracy) on ImageNet-1K image classification.}
    \label{table:diff_sampling}
    \vspace{-1em}
\end{table}

\paragraph{Scaling Vision Models}
Many works~\cite{tan2019efficientnet,riquelme2021moe,zhai2021scaling,swinv2} examine how to scale vision models, but most are more concerned with exploring the perspective of model architecture designs. For example, EfficientNet~\cite{tan2019efficientnet} extensively studied how model width, model depth and input resolution affect the convolutional neural networks; ~\cite{riquelme2021moe} proposed to scale vision model with sparse mixture-of-expert; ~\cite{zhai2021scaling} and ~\cite{swinv2} studied how to scale ViT and Swin Transformer, respectively. 

Only a few works explored the perspective of data scaling under the pre-training fine-tuning paradigm. BiT~\cite{kolesnikov2020big} revisited the supervised pre-training on a wide range of data scales up to 1M images. SEER~\cite{goyal2021self} studied the effectiveness of data scaling in the contrastive learning framework with up to one billion images. Recently, SplitMask~\cite{el2021large} find that masked image modeling is robust to the size of pre-training data and challenges the data scaling capability of masked image modeling, which is most relevant to our work.

\section{Conclusion}
In our work, we systematically study the data scaling capability of masked image modeling at different model sizes and training lengths. Based on the extensive experiments, we demonstrate that masked image modeling is not only a model scalable learner but also a data scalable learner, which challenges the conclusion of previous literature that a large dataset may not be necessary in masked image modeling. The reason behind this is that they overlooked a key factor, namely training length. In addition, a strong correlation between the validation loss of masked image modeling and the fine-tuning performance is observed. This observation suggests that validation loss can be considered as a good proxy metric for evaluating pre-trained models, and makes it possible to reduce the experimental overhead of measuring models by fine-tuning.

While these findings deepen our understanding of masked image modeling in data scaling angles and can facilitate future research, our study still has limitations. First, the maximum model size used in our study reaches only one billion parameters, which we speculate leaves the overfitting phenomenon on the ImageNet-1K dataset unobserved; Second, there is a lack of research on the effect of encoder specifications (\emph{e.g.}, depth and width) on data scaling. Third, our study does not involve the study angle of data augmentation which is a common technique to alleviate data scarcity and overfitting.

\bibliographystyle{apalike}
\bibliography{ref}

\appendix

\section{Hyper-parameters and training details}
We illustrate the training details of pre-training and fine-tuning for different tasks and different models. Table~\ref{table:setting-pretrain} presents pre-training details. Table~\ref{table:setting-finetune} presents the fine-tuning details on ImageNet-1K image classification. Table~\ref{table:setting-inat} presents the fine-tuning details on iNaturalist 2018. Table~\ref{table:setting-coco} presents the fine-tuning details on COCO dataset. Table~\ref{table:setting-ade} presents the fine-tuning details on ADE20K dataset. 

\section{Training dynamics of masked image modeling}
\begin{figure}[htbp!]
    \centering
    \includegraphics[width=1\linewidth]{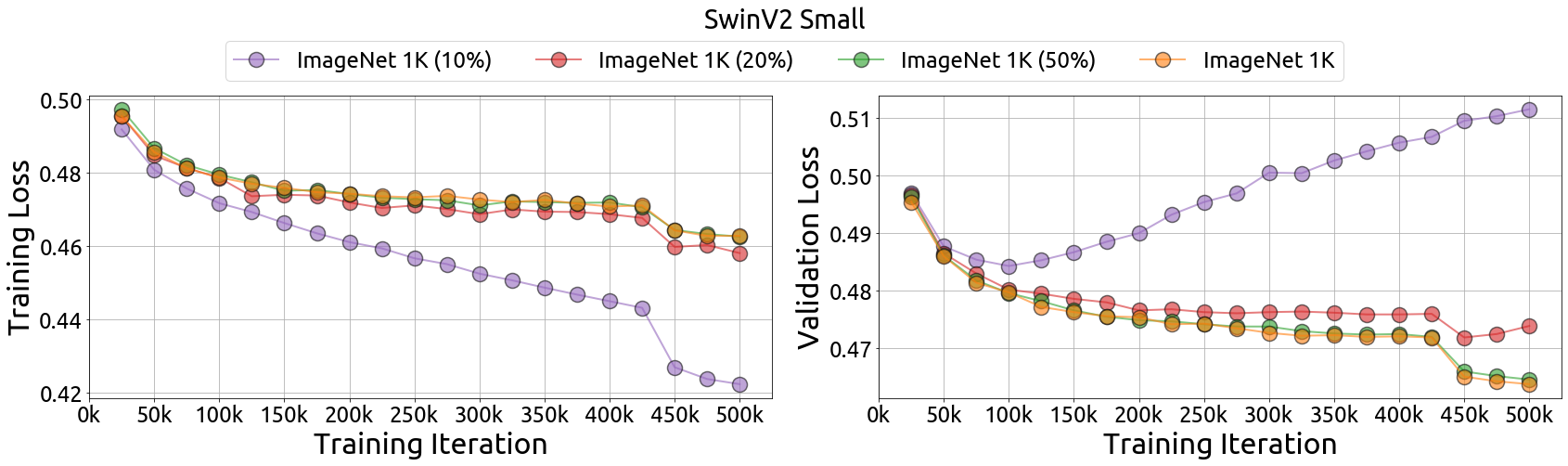}
    \includegraphics[width=1\linewidth]{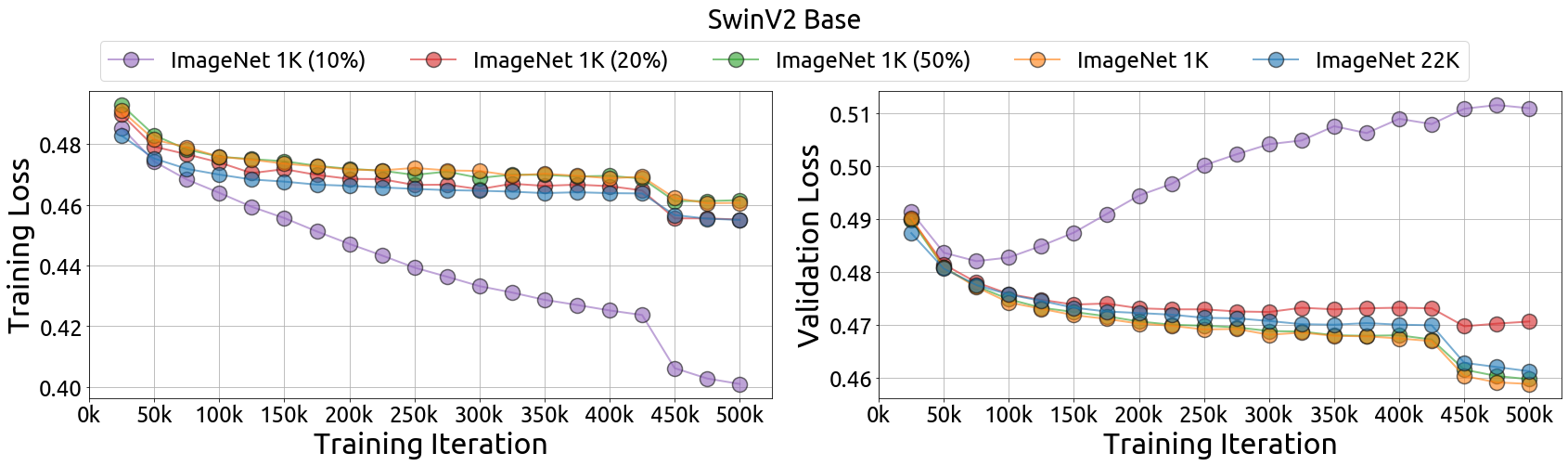}
    \includegraphics[width=1\linewidth]{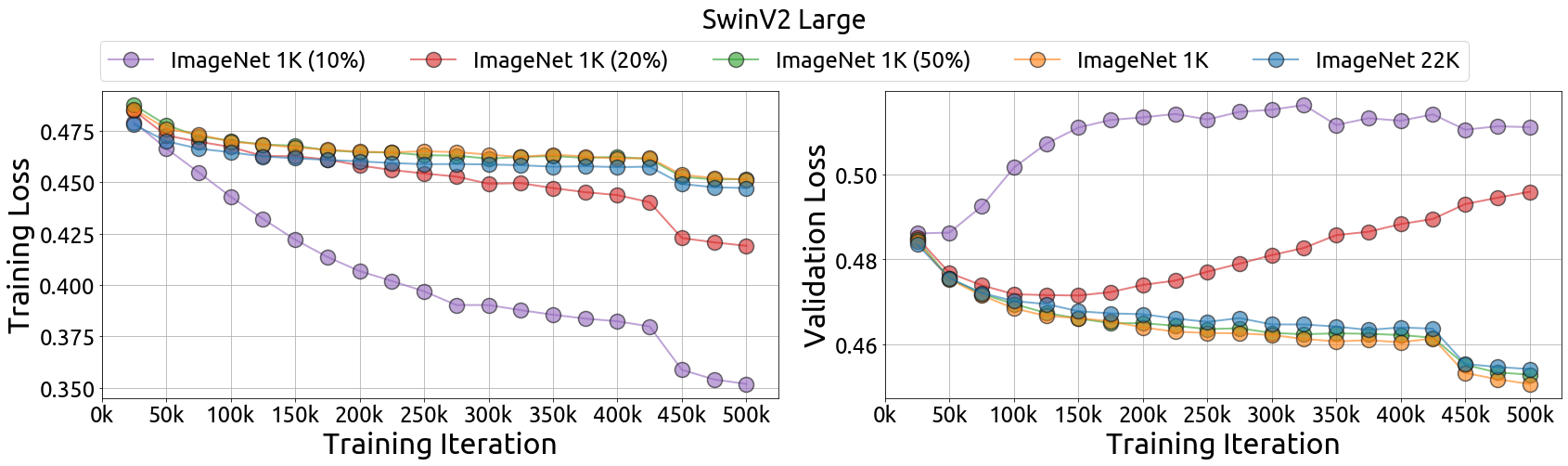}
    \includegraphics[width=1\linewidth]{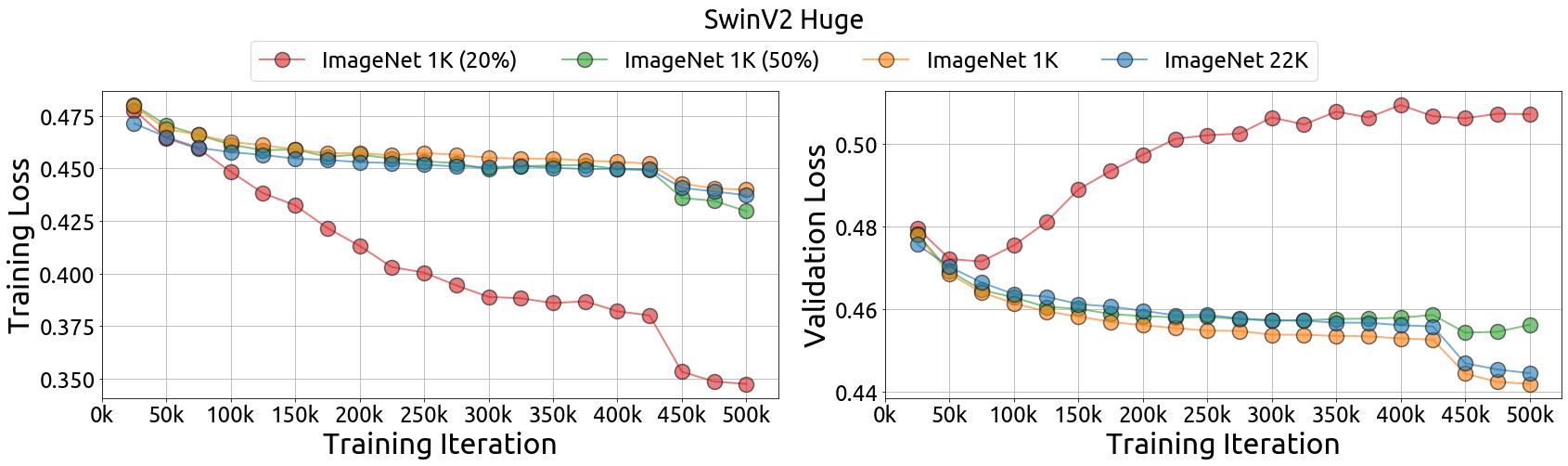}
    \includegraphics[width=1\linewidth]{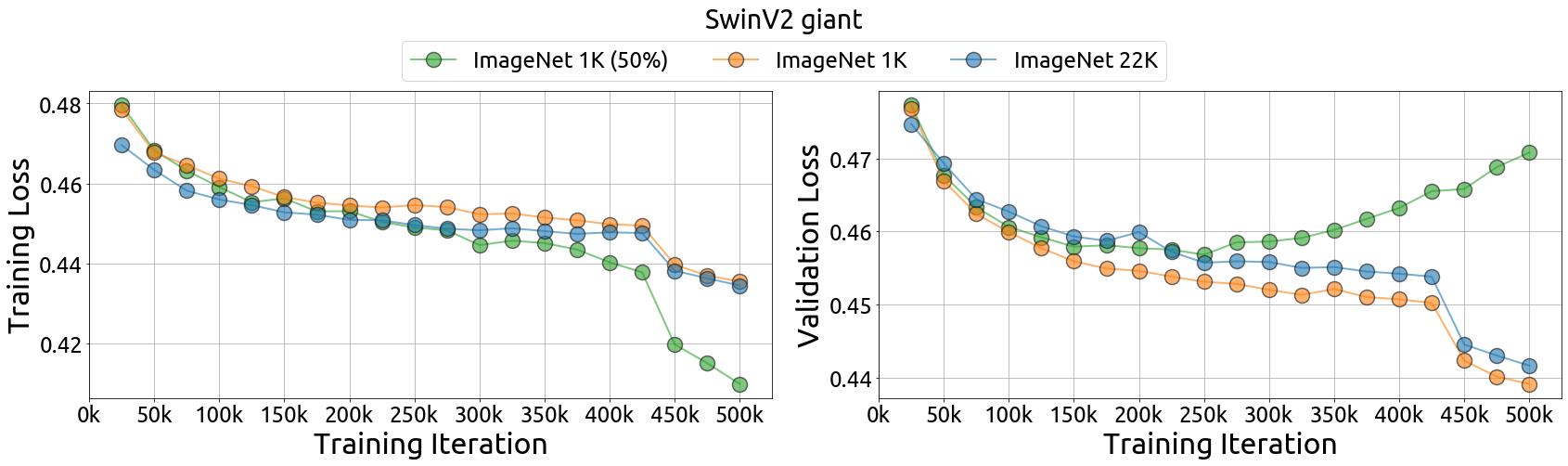}
    \caption{
    Each row presents the training and the validation loss curves for training with the same model (e.g., SwinV2 giant at the last row) but different datasets. The training loss is computed on its corresponding training dataset, and the validation loss is computed on the ImageNet-1K validation set. \emph{Best viewed in color.}
    }
    \label{fig:train_val_loss_diff_model}
\end{figure}

\begin{figure}[htbp!]
    \centering
    \includegraphics[width=1\linewidth]{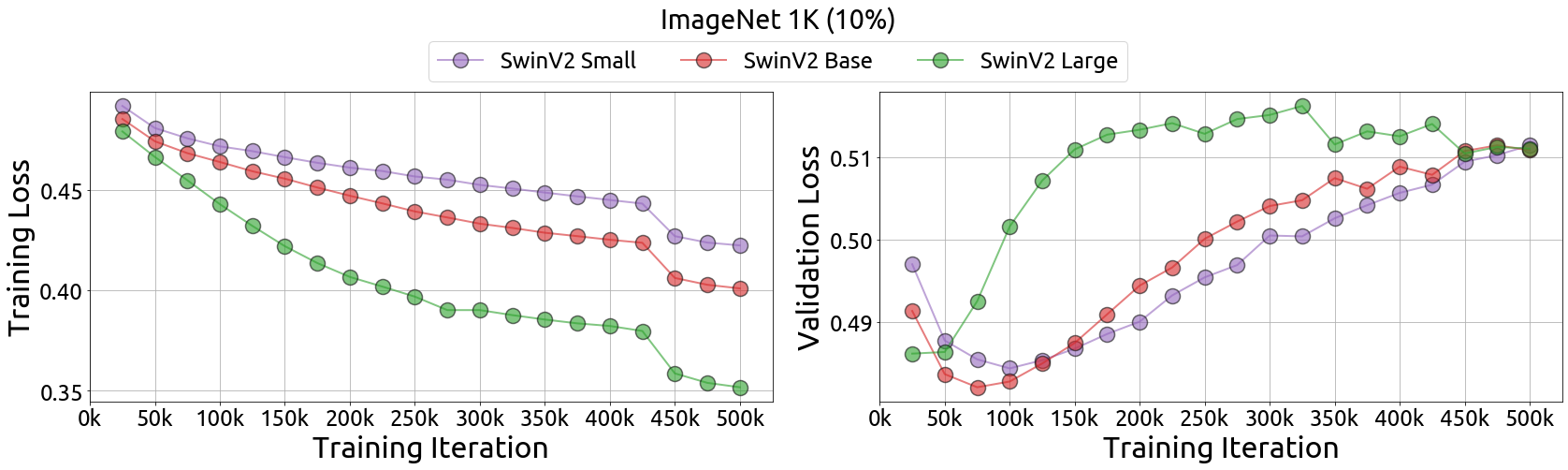}
    \includegraphics[width=1\linewidth]{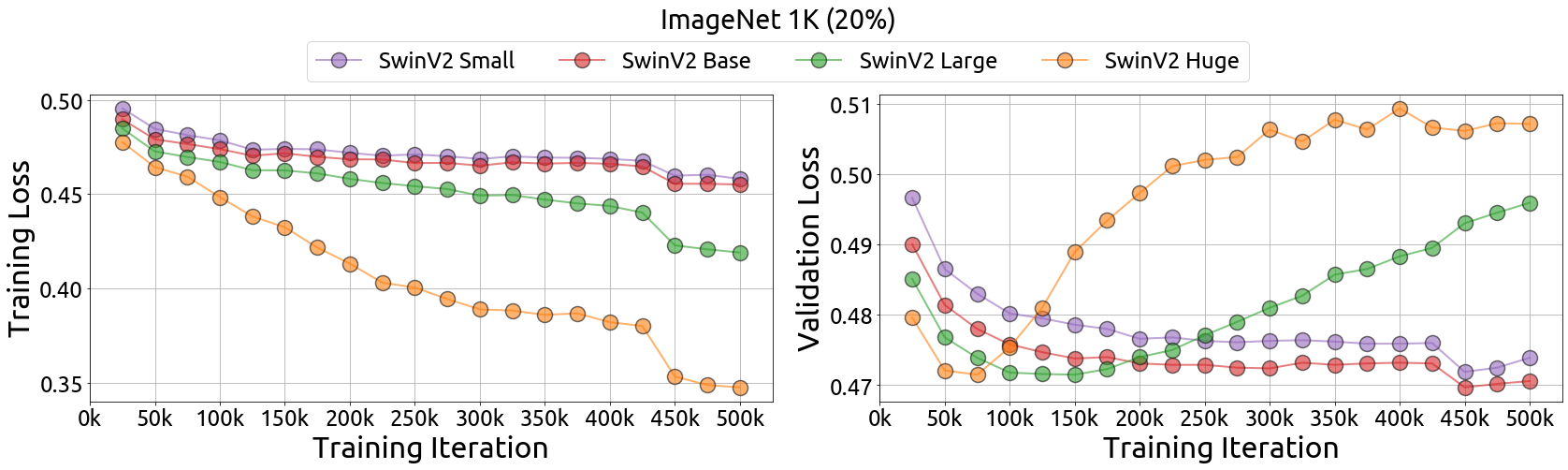}
    \includegraphics[width=1\linewidth]{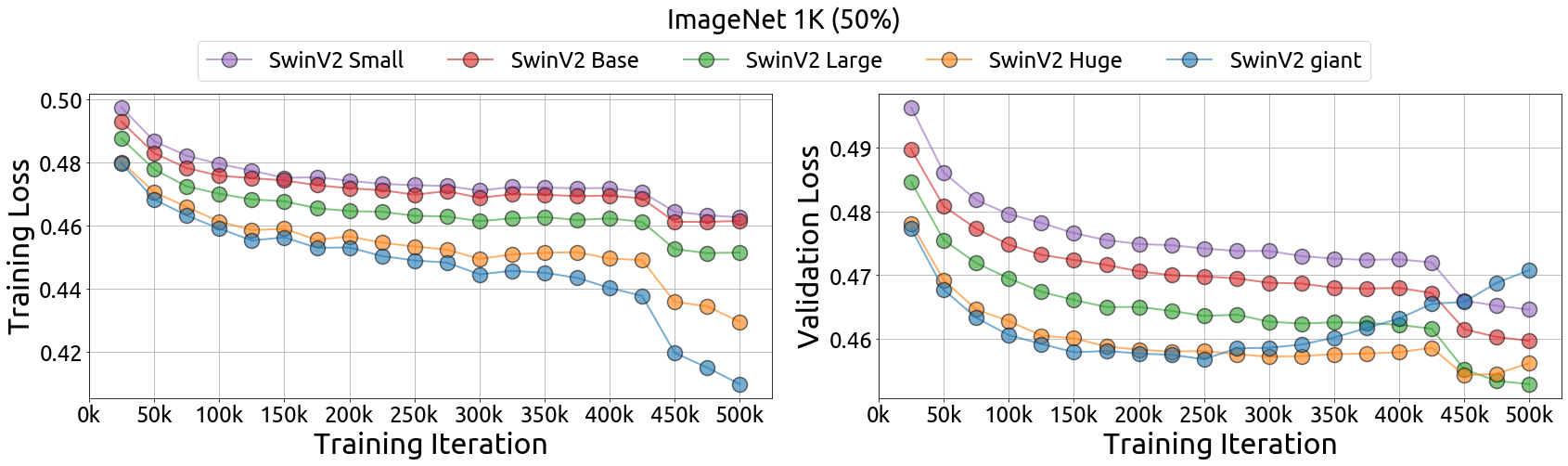}
    \includegraphics[width=1\linewidth]{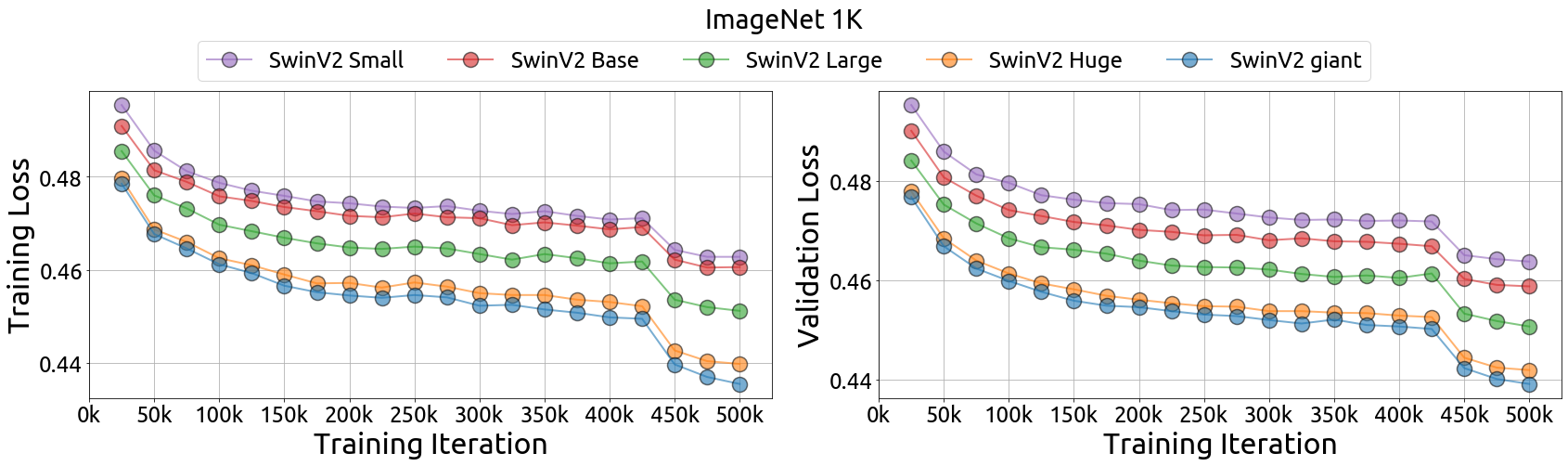}
    \includegraphics[width=1\linewidth]{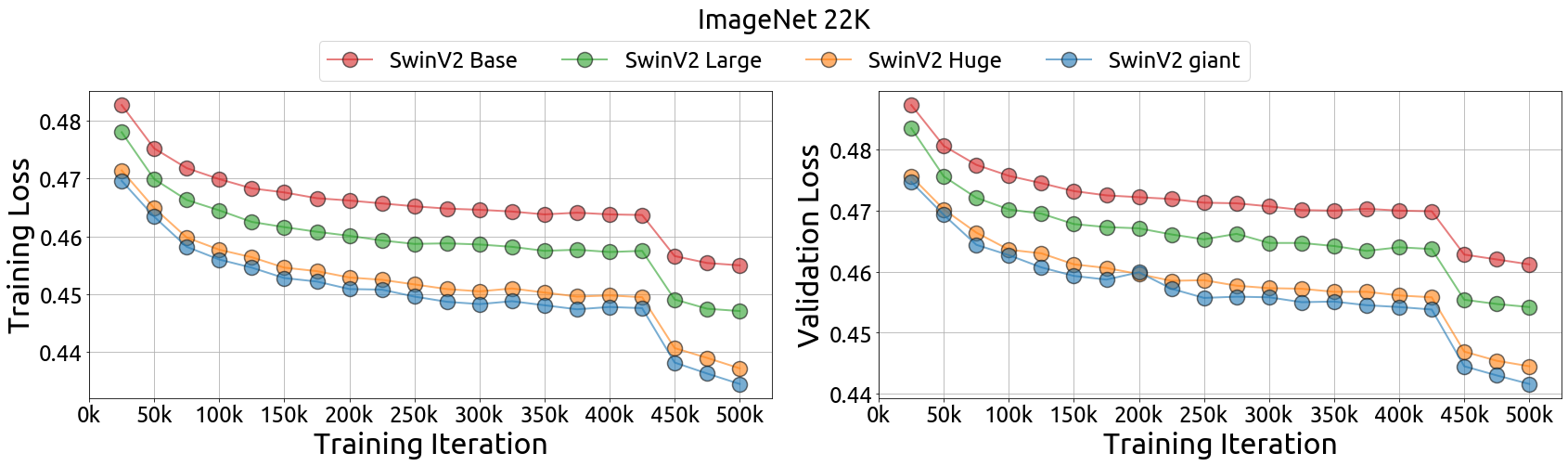}
    \caption{Each row presents the training and the validation loss curves for training with the same dataset (e.g., ImageNet22K at the last row) but different models. The training loss is computed on its corresponding training dataset, and the validation loss is computed on the ImageNet-1K validation set. \emph{Best viewed in color.}}
    \label{fig:train_val_loss_diff_data}
\end{figure}

We show the training curves and validation curves of different models trained by masked image modeling to better illustrate the training dynamics. In Figure~\ref{fig:train_val_loss_diff_model}, each row presents the training and validation loss curves for training with the same model but different dataset. The training loss is computed on its corresponding training dataset and the validation loss is computed on the ImageNet-1K validation set. We make the following observations: First, all models have the overfitting issues when using small datasets. Second, for the non-overfitting cases, the training and validation losses are similar using different sizes of datasets for training.
In Figure~\ref{fig:train_val_loss_diff_data}, the training/validation loss curves of different models but using the same training dataset are presented at each row. We make the following observations: First, larger models have lower training losses than smaller models for all datasets. Second, the validation loss of the larger model is lower than the smaller model in the non-overfitting cases but higher than the smaller model in the over-fitting cases.

\begin{table}[htbp!]\footnotesize
    \centering
    \begin{tabular}{c|c}
    \toprule
    \multicolumn{2}{c}{\textbf{Pre-training setting of all models}} \\
    \hline
    Input size & $192^2$ \\
    Window size & 12 \\
    Patch size & 4 \\
    \hline
    Mask patch size & 32 \\
    Mask ratio & 0.6 \\
    \hline
    Training iterations &  125,000 / 250,000 / 500,000 \\
    Batch size & 2048 \\
    Optimizer & AdamW \\
    Init. learning rate & 4e-4 \\
    Weight decay & 0.05 \\
    Adam $\epsilon$ & 1e-8 \\
    Adam $\beta$ & (0.9, 0.999) \\
    Learning rate scheduler & Step \\
    Step learning rate ratio & 0.1 \\
    Step iterations & 109,375 / 218,750 / 437,500 \\
    Warm-up iterations & 6250 \\
    \hline
    Gradient clipping & 5.0\\
    Stochastic depth & 0.1 \\
    \hline
    Rand crop scale & [0.67, 1] \\
    Rand resize ratio & [3/4, 4/3] \\
    Rand horizontal flip & 0.5 \\
    Reconstruction target & Norm. with sliding window~\cite{fang2022corrupted} \\
    Norm. patch size & 47 \\
    \bottomrule
    \end{tabular}
    \vspace{0.5em}
    \caption{Details and hyper-parameters for SimMIM pre-training.}
    \label{table:setting-pretrain}
\end{table}

\begin{table}[htbp!]\footnotesize
    \centering
    \begin{tabular}{c|ccccc}
    \toprule
    \multirow{2}{*}{\textbf{Hyperparameters}} & \multicolumn{5}{c}{\textbf{SwinV2}} \\
    & \textbf{Small(S)} & \textbf{Base(B)} & \textbf{Large(L)} & \textbf{Huge(H)} & \textbf{giant(g)} \\
    \hline
    Input size & \multicolumn{5}{c}{$224^2$} \\
    Window size & \multicolumn{5}{c}{14} \\
    Patch size & \multicolumn{5}{c}{4} \\
    \hline
    Training epochs & 100 & 100 & 100 & 50 & 50 \\
    Warm-up epochs & 20 & 20 & 20 & 10 & 10 \\
    Layer decay & 0.8 & 0.75 & 0.7 & 0.65 & 0.65 \\
    Batch size & \multicolumn{5}{c}{2048} \\
    Optimizer & \multicolumn{5}{c}{AdamW} \\
    Base learning rate & \multicolumn{5}{c}{5e-3} \\
    Weight decay & \multicolumn{5}{c}{0.05} \\
    Adam $\epsilon$ & \multicolumn{5}{c}{1e-8} \\
    Adam $\beta$ & \multicolumn{5}{c}{(0.9, 0.999)} \\
    Learning rate scheduler & \multicolumn{5}{c}{cosine} \\
    \hline
    Gradient clipping & \multicolumn{5}{c}{5.0}\\
    Stochastic depth & \multicolumn{5}{c}{0.2} \\
    Label smoothing & \multicolumn{5}{c}{0.1} \\
    \hline
    Rand crop scale & \multicolumn{5}{c}{[0.08, 1]} \\
    Rand resize ratio & \multicolumn{5}{c}{[3/4, 4/3]} \\
    Rand horizontal flip & \multicolumn{5}{c}{0.5} \\
    Color jitter & \multicolumn{5}{c}{0.4} \\
    Rand augment & \multicolumn{5}{c}{9 / 0.5} \\
    Rand erasing prob. & \multicolumn{5}{c}{0.25} \\
    Mixup prob. & \multicolumn{5}{c}{0.8} \\
    Cutmix prob. & \multicolumn{5}{c}{1.0} \\
    \bottomrule
    \end{tabular}
    \vspace{0.5em}
    \caption{Details and hyper-parameters for ImageNet-1K fine-tuning.}
    \label{table:setting-finetune}
\end{table}

\begin{table}\footnotesize
    \centering
    \begin{tabular}{c|ccc}
    \toprule
    \multirow{2}{*}{\textbf{Hyperparameters}} & \multicolumn{3}{c}{\textbf{SwinV2}} \\
    & \textbf{Small(S)} & \textbf{Base(B)} & \textbf{Large(L)} \\
    \hline
    Input size & \multicolumn{3}{c}{$224^2$} \\
    Window size & \multicolumn{3}{c}{14} \\
    Patch size & \multicolumn{3}{c}{4} \\
    \hline
    Training epochs & \multicolumn{3}{c}{100} \\
    Warm-up epochs & \multicolumn{3}{c}{20} \\
    Layer decay & 0.8 & 0.75 & 0.7 \\
    Batch size & \multicolumn{3}{c}{2048} \\
    Optimizer & \multicolumn{3}{c}{AdamW} \\
    Base learning rate & \multicolumn{3}{c}{1.6e-2} \\
    Weight decay & \multicolumn{3}{c}{0.1} \\
    Adam $\epsilon$ & \multicolumn{3}{c}{1e-8} \\
    Adam $\beta$ & \multicolumn{3}{c}{(0.9, 0.999)} \\
    Learning rate scheduler & \multicolumn{3}{c}{cosine} \\
    \hline
    Gradient clipping & \multicolumn{3}{c}{5.0}\\
    Stochastic depth & \multicolumn{3}{c}{0.2} \\
    Label smoothing & \multicolumn{3}{c}{0.1} \\
    \hline
    Rand crop scale & \multicolumn{3}{c}{[0.08, 1]} \\
    Rand resize ratio & \multicolumn{3}{c}{[3/4, 4/3]} \\
    Rand horizontal flip & \multicolumn{3}{c}{0.5} \\
    Color jitter & \multicolumn{3}{c}{0.4} \\
    Rand augment & \multicolumn{3}{c}{9 / 0.5} \\
    Rand erasing prob. & \multicolumn{3}{c}{0.25} \\
    Mixup prob. & \multicolumn{3}{c}{0.8} \\
    Cutmix prob. & \multicolumn{3}{c}{1.0} \\
    \bottomrule
    \end{tabular}
    \vspace{0.5em}
    \caption{Details and hyper-parameters for iNaturalist 2018 fine-tuning.}
    \label{table:setting-inat}
\end{table}

\begin{table}[htbp!]\footnotesize
    \centering%
    \begin{tabular}{c|ccc}
    \toprule
    \multirow{2}{*}{\textbf{Hyperparameters}} & \multicolumn{3}{c}{\textbf{SwinV2}} \\
    & \textbf{Small(S)} & \textbf{Base(B)} & \textbf{Large(L)} \\
    \hline
    Detector & \multicolumn{3}{c}{Mask R-CNN} \\
    Window size & \multicolumn{3}{c}{14} \\
    Patch size & \multicolumn{3}{c}{4}\\
    \hline
    Training input size & \multicolumn{3}{c}{(1024, 1024)} \\
    Testing input size & \multicolumn{3}{c}{(800, 1333)} \\
    \hline
    Training epochs & \multicolumn{3}{c}{36} \\
    Warm-up iterations & \multicolumn{3}{c}{500} \\
    Batch size & \multicolumn{3}{c}{32} \\
    Optimizer & \multicolumn{3}{c}{AdamW} \\
    Base learning rate & \multicolumn{3}{c}{8e-5} \\
    Weight decay & \multicolumn{3}{c}{0.05} \\
    Adam $\epsilon$ & \multicolumn{3}{c}{1e-8} \\
    Adam $\beta$ & \multicolumn{3}{c}{(0.9, 0.999)} \\
    Learning rate scheduler & \multicolumn{3}{c}{Step} \\
    Step learning rate ratio & \multicolumn{3}{c}{0.1} \\
    Step epochs & \multicolumn{3}{c}{(27, 33)} \\
    \hline
    Stochastic depth & 0.1 & 0.1 & 0.2 \\
    Rand horizontal flip & \multicolumn{3}{c}{0.5} \\
    Scale Jittering & \multicolumn{3}{c}{[0.1, 2.0]} \\
    \bottomrule
    \end{tabular}
    \vspace{0.5em}
    \caption{Details and hyper-parameters for fine-tuning on the COCO dataset.}
    \label{table:setting-coco}
\end{table}

\begin{table}[htbp!]\footnotesize
    \centering
    \begin{tabular}{c|ccc}
    \toprule
    \multirow{2}{*}{\textbf{Hyperparameters}} & \multicolumn{3}{c}{\textbf{SwinV2}} \\
    & \textbf{Small(S)} & \textbf{Base(B)} & \textbf{Large(L)} \\
    \hline
    Architecture & \multicolumn{3}{c}{UPerNet} \\
    Window size & \multicolumn{3}{c}{20} \\
    Patch size & \multicolumn{3}{c}{4}\\
    \hline
    Training input size & \multicolumn{3}{c}{(640, 640)} \\
    Test input size & \multicolumn{3}{c}{(640, 2560)} \\
    Slide test stride & \multicolumn{3}{c}{(426, 426)} \\
    \hline
    Training iterations & \multicolumn{3}{c}{80,000} \\
    Warm-up iterations & \multicolumn{3}{c}{750} \\
    Layer decay & 0.95 & 0.95 & 0.9 \\
    Batch size & \multicolumn{3}{c}{32} \\
    Optimizer & \multicolumn{3}{c}{AdamW} \\
    Base learning rate & \multicolumn{3}{c}{[1e-4, 3e-4]} \\
    Weight decay & \multicolumn{3}{c}{0.05} \\
    Adam $\epsilon$ & \multicolumn{3}{c}{1e-8} \\
    Adam $\beta$ & \multicolumn{3}{c}{(0.9, 0.999)} \\
    Learning rate scheduler & \multicolumn{3}{c}{Linear} \\
    
    \hline
    Stochastic depth & \multicolumn{3}{c}{0.1} \\
    Rand horizontal flip & \multicolumn{3}{c}{0.5} \\
    Scaling Jittering & \multicolumn{3}{c}{[0.5, 2.0]} \\
    Photo Metric Distortion & \multicolumn{3}{c}{\cmark} \\
    \bottomrule
    \end{tabular}
    \vspace{0.5em}
    \caption{Details and hyper-parameters for fine-tuning on the ADE20K dataset.}
    \label{table:setting-ade}
\end{table}

\end{document}